# From Data Scarcity to Data Care:

# Reimagining Language Technologies for Serbian and other Low-Resource Languages

Smiljana Antonijević Ubois, Ph.D.
*Ainthropology Lab*
ORCID: 0000-0003-0383-4919
smiljana@ainthropology.com

**Abstract**

Large-language models are commonly trained on dominant languages like English, and their representation of low-resource languages typically reflects cultural and linguistic biases present in the source language materials. Using the Serbian language as a case, this study examines the structural, historical, and sociotechnical factors shaping language technology development for low-resource languages in the AI age. Drawing on semi-structured interviews with ten scholars and practitioners—including linguists, digital humanists, and AI developers—it traces challenges rooted in historical destruction of Serbian textual heritage, intensified by contemporary issues that drive reductive, "engineering-first" approaches prioritizing functionality over linguistic nuance. These include superficial transliteration, reliance on English-trained models, data bias, and dataset curation lacking cultural specificity. To address these challenges, the study proposes Data Care—a framework grounded in CARE principles (Collective Benefit, Authority to Control, Responsibility, and Ethics)—that reframes bias mitigation from a post-hoc technical fix to an integral component of corpus design, annotation, and governance, and positions Data Care as a replicable model for building inclusive, sustainable, and culturally grounded language technologies in contexts where traditional LLM development reproduces existing power imbalances and cultural blind spots.

**Introduction**

As we sit down in her office for an interview, mathematics professor Jelena[1] recounts how in 2023 her research team finally secured funding to develop Serbian language models and other language-specific tools from the national government's artificial intelligence (AI) research program. In 2021 and 2022, her team ranked among the top candidates but did not receive funding. "The peer reviewers' explanations were unbelievable to me," Jelena recalls, quoting feedback they received: 'Serbian is a small language, everyone works in English, so why develop language resources for Serbian?'" She notes that the grant competition included dual international and domestic review, and while international reviewers evaluated their application highly, the national committee rated it lower. "To me that was precisely an indication that the need for developing language technologies for Serbian is not recognized. Our national mindset needs to be adjusted so that we understand that it is one of the country's infrastructural projects. Just as we need electricity and internet, we need language technologies," she concludes.

Jelena's experience reflects a broader pattern among researchers consulted in this study, who describe the lack of strategic support for advancing Serbia's linguistic digital ecosystem—conditions typical of low-resource languages (LRLs). LRLs "lack the quantity of data necessary for training statistical and machine learning tools and models" due to limited digital corpora, annotated datasets, small speaker populations, and insufficient computational resources (Liu et al., p. 1). Thus far, large language models (LLMs)—artificial intelligence systems that process and generate human language by analyzing large datasets and identifying statistical patterns—have focused predominantly on widely spoken high-resource languages (HRLs) such as English and Mandarin, leaving LRLs significantly marginalized. Most of the world's 7,000 languages remain technologically under-resourced and underrepresented in AI applications, reinforcing historical inequalities and impacting their digital sustainability, accessibility, and global visibility (Lin et al., 2024; Joshi et al., 2020). This disparity has noticeable consequences: because LLMs are commonly trained on dominant languages like English, their understanding of LRLs often reflects cultural and linguistic biases from those source languages. Recent initiatives have attempted to address these imbalances through multilingual pre-training and cross-lingual transfer, enabling models trained on HRLs to extend some capacity to LRLs (Hu et al., 2020). However, outcomes remain uneven, typically favoring languages linguistically closer to the source language—often English (Joshi et al., 2020; Lauscher et al., 2020; Ponti et al., 2020).

---

[1] All respondent names have been replaced with pseudonyms to ensure their anonymity.

An alternative approach involves developing native language models, sometimes called "sovereign language models" (Bondarenko et al., 2025; Barbereau and Dom, 2024), which are trained on data in a specific language to better reflect local linguistic structures, worldviews, and communicative norms. Examples include GPT-NL for Dutch, SW-3 for Swedish, BgGPT for Bulgarian, and Norwegian GPT. Native models trained from scratch outperform multilingual alternatives for a given language, offering deeper semantic understanding and better handling of morphosyntactic complexity, dialectal variation, and domain-specific terminology (Virtanen et al., 2019). Yet native models demand substantial high-quality training data and computational resources—precisely what LRLs often lack. This creates a fundamental tension— whether to prioritize immediate functionality through multilingual models or invest in slower, resource-intensive work of building native infrastructure that more authentically represents linguistic and cultural specificity. For many LRLs, this choice is not merely technical but deeply political, involving questions of digital sovereignty, cultural preservation, and linguistic justice. Developing effective technologies for underrepresented languages therefore requires more than technical optimization, calling for contextualized ethnographic inquiry into the historical, sociocultural, economic, and technical forces that shape how each LRL enters—or fails to enter—the AI landscape.

This paper takes such an approach through a case study of Serbian, a language characterized by rich dialectal diversity and dual alphabetic systems—Cyrillic and Latin. Drawing on semi-structured interviews with ten scholars and practitioners working on Serbian natural language processing (NLP), this study examines both historical and contemporary barriers to developing language technologies for Serbian. Interviews were conducted in the period March-July 2025, predominantly online, and on average lasted one hour. Participants were recruited through purposive snowball sampling within Serbia's relatively small NLP community to ensure diversity across institutional affiliations, educational backgrounds, and career stages. The sample comprised three mathematicians, two computational linguists, one librarian, one data scientist, one philosopher of language, one software engineer, and one digital humanist. Most interviewees (8/10) were in the age group 35-50, and the majority (6/10) were women.

In what follows, I first outline the historical context of the Serbian language and its current state of digital readiness. I then present interview findings that highlight key barriers and opportunities in advancing Serbian language technologies, with particular attention to corpus construction, institutional support, data bias, and interdisciplinary collaboration. Building on these insights, the concluding section identifies strategic areas for intervention, and proposes Data CARE (Collective Benefit, Authority to Control, Responsibility, and Ethics) as a framework that reframes bias mitigation from a post-hoc technical fix to an integral component of corpus design and inclusive language technology development.

**Serbian Linguistic and Technological Context**

*Historical-Linguistic Context*

      Serbian is an Indo-European language that developed a rich medieval literary tradition, including seminal works such as Miroslav's Gospel (c. 1180), Vukan's Gospel (c. 1202), St. Sava's Nomocanon (1219), and Dušan's Code (1349). However, Ottoman occupation (mid 15th–early 19th centuries) devastated this tradition. Monastic libraries, central to manuscript production, faced destruction and looting, with manuscripts burned, stolen, or sold to collectors in Russia, Austria, or Venice.[2] The collapse of urban centers, coupled with the dismantling of the Serbian royal court, eliminated patronage for literary and artistic works; Ottoman authorities further stifled cultural production by restricting the Church's ability to maintain schools and scriptoria, prioritizing instead Islamic education. Secular cultural production also almost entirely ceased, prompting a shift toward oral traditions as the primary means of preserving national identity and culture. Written cultural efforts persisted in the Serbian diaspora, laying groundwork for cultural revival.

      After liberation from Ottoman occupation (1804–15), Serbia gradually rebuilt its literary and cultural infrastructure, but its textual legacy continued to face setbacks. The Balkan Wars (1912–13) and World War I caused significant destruction, while the most catastrophic loss came in 1941, when German bombs obliterated the National Library of Serbia, destroying over 500,000 volumes, including 1,424 medieval Cyrillic manuscripts (Ristić, 2016).

      These losses left Serbian linguistic and literary resources significantly diminished, with lasting implications for language technologies development, particularly for language models

---

[2] For instance, the *Vukan Gospel* is now in the Russian National Library, while *The Serbian Munich Psalter* had been taken to Bavaria in 1688.

where resource scarcity directly shapes dataset scope and quality. Today, data scarcity stemming from centuries of destruction, suppression, and looting of Serbia's written legacy is compounded by contemporary challenges such as limited digitization, restrictive copyright regime, and insufficient institutional support —issues examined in detail in the following section.

Another historical development with contemporary relevance is Serbian linguistic modernization during post-Ottoman cultural revival. Serbian philologist and ethnologist Vuk Karadžić led a comprehensive language reform in the early 19th century, transforming not only Serbian but also Croatian linguistic standardization. This influenced the future of Serbo-Croatian and the broader South Slavic linguistic landscape, with the Vienna Literary Agreement of 1850— signed by Serbian, Croatian, and Slovenian intellectuals—declaring a shared language with two scripts, Cyrillic and Latin (Greenberg, 2004).This language became official after World War I, when Kingdom of Serbs, Croats and Slovenes was formed in 1918. The country was renamed Yugoslavia in 1929, and the 1921 Constitution declared "Serbo-Croato-Slovene" (later on Serbo-Croatian) the official language. Subsequent Yugoslav constitutions maintained this official status until the language fragmented into nationally framed variants following Yugoslavia's disintegration in the 1990s.

Today, Serbian is classified as a national pluricentric language—one spoken across at least two nations where it serves as a national identity marker, enjoys state-sanctioned status with legally prescribed usage in official domains, and has sufficient speakers to establish its own codifying and standardizing linguistic center (Muhr et al., 2020). The ISO 639-3[3] code "hbs" (Serbo-Croatian) classifies it as a macrolanguage, grouping individual languages that share a common linguistic base but are treated as distinct for cultural, political, or national reasons (Eberhard et al., 2025). Serbian is currently spoken by 10-12 million native speakers worldwide, (with hbs speakers estimated at 17-30 million); it is the official language of the Republic of Serbia and Republic of Srpska, co-official in Bosnia and Herzegovina, Montenegro, and Kosovo/ Kosovo and Metohija under UN resolution 1244[4], and an official minority language in Croatia, North Macedonia, Hungary, Romania, Slovakia, and the Czech Republic .

---

[3] ISO 639-3 is an international standard that extends unique three-letter codes to all known natural languages for identification and documentation.

This historical background bears considerable relevance for contemporary language technologies. Post-Yugoslav linguistic and identity politics have shaped language technology development across the former republics, impeding collaborative efforts to create shared language resources and tools— precisely the kind of pooled resources that would yield larger datasets, critically needed for all lesser-resourced languages in the region. For instance, while several respondents proposed developing language models that integrate Serbian, Croatian, Bosnian, and Montenegrin within the pluricentric hbs macrolanguage—following examples such as BERTić and YugoGPT, and supported by research showing that incorporating related languages into training datasets improves model performance (Wang et al., 2023; Snæbjarnarson et al., 2023)—they also noted that current political and institutional conditions in the post-Yugoslav region tend to favor national models, such as Croatia's HR-LLM (Štefanec et al., 2024), over shared ones.

***State of Digital Readiness***

In 2024, 92.5% of urban and 85.5% of rural households in Serbia possessed a computer and broadband internet, with 87.7% of citizens using the Internet regularly (Republički zavod za statistiku, 2025). Serbia's Information and Communication Technologies (ICT) sector has grown notably in workforce and exports, particularly in software development and AI. Since 2017, the Office for IT and e-Government has operated a National Open Data Portal[5] linked to the European Union's (EU), and an e-Government portal with over 1,000 services. Following the 2020 adoption of its National AI Strategy[6], Serbia launched the Institute for AI Research and Development[7] and a supercomputing Data Center in Kragujevac.[8] In 2024, Serbia chaired the OECD's Global Partnership on AI (GPAI) and hosted its first annual summit. A new National AI Strategy (2024–2030), adopted in 2025, emphasizes job creation, quality of life, and alignment with EU, UNESCO, and Council of Europe standards.[9]

---

[4] A territory that declared independence from Serbia in 2008; not recognized by the United Nations as an independent state.

But despite solid digital readiness and AI focus, Serbian is categorized as fragmentarily-supported language according to the European Language Equality (ELE) classification (Krstev and Stanković, 2023). The ELE framework evaluates digital readiness by assessing technological factors (existing resources like corpora and tools) and contextual factors (demographic vitality, investment, research, education, institutional status), combining them into a Digital-Language-Equality (DLE) metric that benchmarks European languages against English across five support tiers: excellent, good, moderate, fragmentary, or weak/no support (Gaspari et al, 2022). Consistent with ELE's classification of Serbian as fragmentarily supported, most respondents view current language resource development as insufficient but improving, highlighting challenges and opportunities in the AI age, which are discussed in the next section.

---

[5] See: https://hub.data.gov.rs/en/home/
[6] See: https://www.media.srbija.gov.rs/medsrp/dokumenti/strategy_artificial_intelligence.pdf)
[7] See: https://ivi.ac.rs/en/
[8] See: https://dct.rs/en/data-center.html)
[9] See: https://www.srbija.gov.rs/tekst/en/149169/strategy-for-the-development-of-artificial-intelligence-in-the-republic-of-serbia.php

**Benefits and Challenges of Serbian Language Technologies**

Reflecting on Serbian's position in the development of language technologies, Milan, a librarian, notes that its relatively small speaker base and devastated textual corpora have historically placed it at a disadvantage; compounded by insufficient awareness among cultural institutions and policymakers, this environment allowed only incremental progress. Today, however, new perspectives are emerging. Ana, a mathematician and an NLP researcher, observes that awareness of digital language technologies' importance has expanded from a small circle of enthusiasts to multiple groups actively advancing the field—an evolution she characterizes as a "promising upward trajectory."

Interviewees converged on their informal rankings of the most influential institutions developing language technologies in Serbia: the University of Belgrade's Faculty of Electrical Engineering (home to the Group for Language Technologies since 1979), Faculty of Mining and Geology, and Faculty of Philology; the University of Niš Faculty of Electronic Engineering; the Mathematical Institute of the Serbian Academy of Sciences and Arts; and specialized organizations such as the Language Resources and Technologies Society (JERTEH)[10] and the ReLDI Center[11]. Frequently cited projects included "Text Embeddings – Serbian Language Applications"(TESLA)[12], developing open-source pre-trained language models for Serbian, and ComText [13], building resources and tools for both Ekavian and Ijekavian Serbian variants. As the next important step, interviewees identify the development of native LLMs capable of capturing Serbian's linguistic richness and enabling long-term sustainability, sovereignty, and innovation.

---

[10] See: https://jerteh.rs
[11] See: https://reldi.rs
[12] See: https://tesla.rgf.bg.ac.rs/index.php/en/project/
[13] See: https://www.ic.etf.bg.ac.rs/projects/comtext-sr/

Participants presented several arguments in favor of developing native language models trained on Serbian as spoken across the Serbian-speaking regions. Their initial argument aligns with contemporary attention to LLMs as tools for language preservation (Koc, 2025), positing that native LLMs help preserve Serbian in the age of AI. "Just as languages that never developed a writing system gradually declined and eventually disappeared with their last speaker, so too will languages that fail to build digital infrastructure and resources slowly fade away," Jelena observes.

Respondents also emphasize that multilingual LLMs, trained on dozens or even hundreds of languages, cannot substitute for native models because they capture Serbian only superficially through cross-lingual transfer. "You train the model on English, and then it learns how to roughly reproduce that in Serbian, but it lacks nuance, cultural and historical context—it simply doesn't have the breadth or depth," explains Marko, a data scientist. He argues that no cross-lingual transfer can replace native training data as "what isn't in the data won't be in the model." This assessment aligns with research showing that although multilingual models can generalize across languages, native models trained for low-resource languages from scratch systematically outperform them (Virtanen et al., 2019).

This is particularly relevant for Serbian, given its linguistic diversity: two scripts, five dialects, two phonological variants, and unique morphosyntactic features. Vera, a philosopher of language, describes Serbian as rooted in dialectal and ethnic pluralism explaining that in the province of Vojvodina alone the same Serbian dialect connects over 20 ethnic communities, each contributing something of their own. "That level of diversity is not the case for many other European languages that are often far more ethnically monolithic," she points out. Extending this argument, Milan questions how models can meaningfully capture Serbia's history and culture without training on Serbian texts, relying instead on data sourced from other languages. This aligns with Windsor's (2022) argument on the role of language technologies and data in understanding international relations (IR): "Because non-English corpora are underrepresented in IR research using text-as-data methods, scholars' understanding of international political phenomena is filtered through an ethnocentric, Western lens. This can lead to scholars misdiagnosing or failing to understand political phenomena since the experiences of political processes omit the perspectives offered by accounts in non-English languages" (p.2).

Economic and security considerations also figure prominently in arguments for native language models. Respondents note that commercial LLMs are not only costly, but also unfeasible for the public sector. Concerns about data sovereignty, security, and compliance require that any model used in the public sector be domestically trained and hosted, without foreign infrastructural dependence. Beyond such practical arguments, several respondents emphasized the affective and symbolic dimensions of native language models development. "You feel left out when your language isn't represented," Ana observes; "culturally and linguistically, it's an alienating experience." She points out that this sense of exclusion is especially stark given current European strategies to build LLMs in support of a European language family. "Without an LLM for Serbian we are effectively left out of that family, even though we belong to it," she concluded. This exclusion, respondents warned, is not just technological—it is cultural.

Yet, developing LLMs depends on large, high-quality corpora—an area where Serbia continues to face critical gaps due to insufficient data and the lack of systematic efforts, topics to which we turn next.

*Limited corpora and institutional support*

A corpus (plural: corpora) is a large, structured collection of authentic language data—written or spoken—presented in machine-readable format, purposefully selected and sampled to be representative of a specific language, and often annotated with linguistic metadata (McEnery, Xiao, & Tono, 2006, p. 5). Large, high-quality, and representative corpora provide the statistical foundation for computational models to learn linguistic patterns, structures, and nuances necessary to accurately understand and generate language. However, the lack of adequate corpora remains one of Serbia's most significant challenges, rooted in both historical and contemporary factors.

The severity of this problem comes up immediately in practitioners' accounts. "The problem with data is that it simply doesn't exist," states Nikola, a mathematician. Vera concurs: "When it comes to language data, there really isn't enough of it—that's a fact." Dragana, a computational linguist, reframes the challenge: "The problem lies in how to create adequate datasets. In general, there isn't as much material available [for Serbian] as there is for some other languages."[14]

Interviewees explain that most training data for Serbian language models currently come from scraped web sources—material they describe as low-quality and biased toward the most prevalent online content, such as entertainment, casual conversation, and daily news. Ana articulates the fundamental problem: "If we are pulling content from the web, how can we evaluate what portion is of sufficient quality for inclusion in a dataset, what portion is redundant, and what portion contains low-quality language that might later negatively impact model behavior — things like bias or unexpected outputs?"

---

[14] For a detailed overview of existing language models and datasets for Serbian, see Škorić (2024).

And even with scraping web materials, the lack of corpora remains a pressing problem. Nikola illustrates this with the case of "Kišobran" [Umbrella][15], the largest Serbian-language corpus available on Hugging Face. This project aggregates and de-duplicates data from sources like Common Crawl and Archive.org, yielding approximately 23 billion words. However, this represents not just Serbian, but a combined corpus of South Slavic languages (former Serbo-Croatian). Serbian alone accounts for 7 to 9 billion words; by contrast, English-language models train on between 3 and 14 trillion words. "The gap in data volume is staggering," Nikola concludes.

Rather than relying solely on web-scraped materials, interviewees advocate for a corpus covering the full spectrum of language use: spoken and written forms, literature, television broadcasts, newspapers, public speeches, educational and academic discourse, and domain-specific terminology across fields such as law and medicine. As Sandra, a digital humanist, explainss, such datasets should encompass "everything that represents the broader Serbian cultural, scientific, and other contexts"—materials of national importance. But building such comprehensive, high-quality datasets remains hindered by limited digitization and restrictive copyright.

Respondents acknowledge steady progress in digitization since the 1980s, especially over the past two decades, yet identify several systemic problems. One is the fragmented and uncoordinated nature of present-day digitization, which leads to duplication of work across institutions while leaving significant gaps. Another is the "early-digitization error," when materials were scanned as images without OCR, rendering them non-searchable and unusable for language-technology projects today. This omission was not unique to Serbia; early digitization efforts elsewhere often omitted OCR due to limited accuracy, cost constraints, and a preservation philosophy that viewed digitization as faithful reproduction rather than textual transformation. Today, "the challenge lies in converting and transforming those images into actual text, especially when we encounter the specific complexities of the [Serbian] language, such as the use of both Cyrillic and Latin scripts," Ana notes.

---

[15] See: https://huggingface.co/datasets/procesaur/kisobran

Copyright restrictions further intensify challenges of inadequate datasets by limiting access to both born-digital and digitized materials. "High-quality literature, which represents beautiful language, is ideal for language models. However, copyright restrictions make this just a dream," Ana underscores. When asked why a larger high-quality corpus for Serbian does not exist, Dragana responds simply: "Copyright." Marta echoes this frustration, stating she has "no sympathy for copyright" because "any work—scientific, artistic, whatever it may be—is a public good when it concerns language." Scholarship on low-resource languages confirms this problem: copyright restrictions exacerbate data scarcity by limiting access to precisely the kind of materials LRLs need to build comprehensive, high-quality language models (Moody, 2023; Siminyu et al., 2021). Certain countries, such as Wales and Iceland, have addressed this by working with writers' associations and publishers to secure copyright owners' permissions for using materials free of charge (Helgadóttir et al., 2012; Knight at al., 2020). Serbia's copyright law aligns with EU standards but lacks an AI-training exception comparable to EU Directive 2019/790, which allows text and data mining for AI training unless rights-holders have opted out.

Alongside legal harmonization, Serbia's language technology development also involves EU project funding. However, this funding primarily supports multilingual resources, leaving national language development to individual states. Respondents argue, however, that Serbian institutions tasked with advancing language technologies have shown limited engagement. "Even for basic tools like Serbian spell-check or a script converter from Latin to Cyrillic, there isn't a single authoritative tool produced by these institutions," Marko observes.

This limited institutional engagement proves particularly consequential given that academic institutions lack capacity to lead large-scale national projects and the private sector provides no alternative. Most Serbian IT firms focus on international markets with no commercial incentive to develop Serbian-specific tools, respondents note, while domestic-facing companies—primarily serving public administration—remain contract-driven and lack resources for independent research and development. "This is a clear-cut case for state intervention," Marko concludes.

But such intervention remains weak, respondents argue, despite Serbia's latest (2025) national AI strategy including language technologies. "It still hasn't reached the level of national understanding that this [language technology development] requires significant investment," Dragana explains. Vera elaborates: "There's pressure and expectations from the [political] top to be highly efficient in digitally transforming society, but in real life, it's all extremely difficult and barely feasible given the lack of resources. It's as if we expect results to just appear out of thin air."

This critique extends to the state-funded Institute for Artificial Intelligence, which, according to respondents, has not prioritized authentic Serbian-language development. Marta illustrates: "They didn't use texts written in Serbian. They took existing OpenAI content, machine-translated it into Serbian, and then used that as training material. That's why the model generates made-up words like *аффекција*."[16] For respondents, this signals not only a technical failure but also a missed opportunity: instead of building technologies grounded in Serbia's linguistic and cultural realities based on authentic materials, state-sponsored institutions relied on machine translations yielding inaccurate data susceptible to data bias—the very issue explored in the following section.

**Data Bias**

Data bias in language technologies is not a mere technical flaw, but a reflection of broader societal and epistemological issues. Datasets reflect what is present in a society, reproducing and amplifying socio-cultural, political, racial, economic, and gender-based biases. This stems from multiple sources: limited data availability, underrepresentation of certain groups and topics, subjective annotation practices, and unexamined assumptions in corpus design and selection (Bender et al., 2021; Blodgett et al., 2020; Gebru et al., 2021). Although bias may sometimes result from the inherent structure of data—such as gendered disease patterns in medicine—interviewees underline that its primary source lies in how datasets are created and curated. "Data bias means lack of representativeness—the model gives answers based on what it has seen most often, frequently favoring particular groups, phenomena, or activities," Dragana explains. "That doesn't mean it's true, or even the majority opinion—it's just what the model has encountered most frequently." She recounts prompting a commercial language model for jokes about Serbs and receiving "a stream of awful, deeply offensive content." Respondents stress that such outputs are not anomalies but structural symptoms of data imbalances—of what the model has and, crucially, has *not* been exposed to during training.

---

[16] Transliteration-based hallucination of the English word affection.

Data bias is particularly pronounced in low-resource languages, where data scarcity combined with cross-lingual transfer from HRLs produces models that underperform and reinforce existing representational gaps. Multilingual models trained predominantly on HRLs tend to encode and transfer those dominant languages' linguistic, cultural, and social biases resulting in misclassification of named entities, gendered translation errors, and overall degraded performance. This creates a double disadvantage for LRLs—underrepresentation combined with distorted representations rooted in unrelated linguistic contexts.

In LRLs, data bias is often amplified by the lack of proper data annotation—often invisible labor that requires prolonged intellectual and manual work and adequate knowledge. Scarcity of qualified and motivated workforce for such efforts is yet another setback in Serbia: "The linguistic community is small, especially when it comes to computational linguists," Marta explains, while Dragana adds that "there aren't that many people interested in building the foundational resources; AI sounds fashionable but few actually want to do that groundwork."

Petar, a software engineer, adds that proper annotation must be paired with rigorous evaluation, since language models are inherently imperfect and inevitably contain some error. He critiques the historical reliance on quantitative metrics—"poor proxies for model performance," as he describes them— and welcomes the recent shift toward more nuanced evaluation criteria, including human expectations, creativity, linguistic quality, fairness, and sensitivity. "The field is finally engaging with these questions in a meaningful way," Peter concludes, pointing to a growing awareness of the multidimensional nature of model outputs.

It is important to note that high-resource languages are not immune to data bias either. As Ana notes: "Not even the British national corpus is an adequate, fully representative corpus. It primarily reflects the language of the middle class, adult males." This illustrates that bias is not merely a function of data volume but of whose voices are systematically included or excluded. Developing more equitable language technologies thus requires sustained attention to the sociopolitical contexts of data creation, not just improvements in technical performance.

In that regard, interviewees voice concerns that framing bias as an abstract algorithmic problem distracts from its socio-cultural origins. "We're implicitly blaming some abstract entity — an algorithm or dataset — instead of focusing on how datasets are created," Vera explains. Rather, shared human accountability across all stages of AI development is needed, as responsibility for mitigating data bias is necessarily distributed. "We're all responsible [for data bias]—from those who collect and label data to those who design the systems," Ana argues. Petar adds that responsibility also lies with the deployer — the one who places the model in front of users and must ensure its behavior aligns with intended purposes. Given this broad responsibility for ensuring representational justice, trustworthiness, and inclusiveness in AI systems, participants emphasize the importance of interdisciplinary collaboration—a theme the following section examines in detail.

**Interdisciplinary Collaboration**

Among interviewed experts, interdisciplinary collaboration is common and valued in language technology development, with teams typically composed of computational and general linguists, mathematicians, computer scientists, librarians, and domain experts. Yet, respondents stress the persistence of two intellectual traditions: the engineering perspective, which often treats language as just another type of data and emphasizes efficiency, and the humanistic perspective, which views language as socially and culturally embedded.

On this subject, interviewees recounted scenarios where programmers, often operating under pressure to deliver results, underestimated the complexity of linguistic or cultural representation. For instance, Vera recalls advocating for greater inclusion of women authors in the training corpus—a move grounded in principles of diversity and representational justice—only to find some teammates prioritizing feasibility within the available computational and financial resources. Interviewees thus accentuate that humanities and social science (HSS) research contributes not only through ethical oversight but also methodologically, by questioning the assumptions guiding model design. While engineers may prioritize performance and functions, philosophers can unpack the epistemological and ethical implications of technical choices, identifying which choices carry real-world consequences beyond legal compliance.

However, HSS scholars' ability to effectively contribute to language technology development and assessment presupposes foundational understanding of how these systems work, particularly their inherent limitations and typical error patterns. As Ana remarks, foundational technical literacy—such as recognizing the kinds of mistakes statistical models tend to produce— is indispensable for effective engagement. Peter notes that HSS scholars provide invaluable frameworks for interrogating what gets encoded in training data and how certain populations may be included or excluded in computational representations, but this critique is not effective without technical grounding, necessary for high-quality contributions to AI ethics. While interviewees do not expect HSS scholars to become data scientists or software engineers, they maintain that working knowledge of data structures and machine learning workflows can significantly enhance their impact. For instance, Marta emphasizes that annotating corpora and evaluating tagged data requires an understanding of both linguistic and computational constraints, especially when working with large datasets. Marko reinforces this by arguing that concepts such as tokenization, attention mechanisms, or hallucination should be part of the conceptual toolkit of digital humanists and computational social scientists.

Respondents note that effective models of knowledge integration are found in longstanding interdisciplinary initiatives, such as JERTEH, and in more recent efforts where interdisciplinary collaborations become genuinely co-constructive over time—programmers gain more nuanced appreciation of linguistic complexity, while linguists benefit from exposure to technical constraints and problem-solving approaches. To illustrate, Dragana observes that although "programmers often know very little about language and think certain things can be done easily," such assumptions, though naïve, can spark creative solutions valuable in the early stages of technology development.

Ultimately, more robust collaboration between HSS and technical domains requires educational reform, interdisciplinary training, and institutional recognition of digital humanities and critical AI studies as necessary fields. Without this, the divide between social critique and technological design risks producing language technologies that fail to reflect the values, diversity, and realities of the societies they are meant to serve. As Milan warns, if HSS scholars cannot explain their relevance to the broader public—or to engineers—they risk becoming sidelined in one of the most consequential technological developments of our time.

**Towards Caring for Low-Resource Languages**

Developing language technologies for low-resource languages like Serbian presents both significant opportunities and persistent challenges. Employing a contextualized ethnographic approach, this study situates these challenges within a broader struggle for knowledge equity, digital justice, and linguistic inclusion. Insights from in-depth interviews with Serbian scholars and practitioners enable close examination of the structural factors shaping language technology development in the AI age. Drawing on these accounts, this section synthesizes key findings and outlines strategic interventions that address both historical and contemporary barriers to building NLP capacity for LRLs such as Serbian.

One of the most significant insights concerns the inherent limitations of cross-lingual transfer in multilingual LLMs and the importance of native language models. Approaches relying on machine-translated data or localized versions of multilingual LLMs—optimized primarily for dominant languages—reveal deep methodological and linguistic weaknesses. These methods provide only shallow representations of LRLs, risking culturally inauthentic, nonsensical, or even offensive outputs that reproduce bias and reinforce linguistic marginalization. Studies confirm that native LRL models trained on diverse, authentic data surpass multilingual alternatives, demonstrating the irreplaceable value of language-specific training corpora over cross-lingual transfer.

Beyond technical performance, native models carry broader cultural significance by preserving languages and linguistic heritage in the AI age, enhancing cultural visibility and addressing the alienation citizens feel when their mother tongue is excluded from the global AI arena. Native models also address economic and security concerns: public-sector applications require domestically trained and hosted models to ensure data sovereignty, regulatory compliance, and independence from foreign and/or commercial providers. This study thus strongly supports respondents' plea for treating language technologies as core infrastructure, with sustained investment in curated, representative, open-access corpora and native models developed in the public interest.

Another key insight concerns data scarcity, which Serbia exemplifies acutely: much of the available training data is scraped from the web, where quality is uneven and can degrade model behavior, producing inaccurate or biased outputs. Rooted in historical disruptions, this scarcity is further deepened by limited digitization and restrictive copyright requirements. To address these challenges, successful digitization initiatives provide actionable models for Serbian and other low-resource languages, demonstrating how coordinated high-quality scanning, robust metadata, and open-access policies can build representative repositories that connect libraries, archives, and cultural institutions into a unified digital ecosystem (NARA, 2014, 2023; United Nations, 2020). Likewise, to navigate copyright constraints, Serbia and other LRL contexts could adopt opt-in frameworks similar to those used in Iceland and Wales, encouraging authors to contribute their materials voluntarily and thus accelerating the creation of high-quality datasets. Collectively, these approaches outline concrete pathways for addressing systemic challenges of LRLs while promoting linguistic diversity and digital inclusion.

This study also demonstrates that LRLs face distinctive forms of data bias, including representational bias (skew toward low-quality web content), amplification bias (the reinforcement of existing social imbalances, such as gender disparities in available texts), and annotation bias (arising from limited expert workforce in small linguistic communities). In response, I propose Data Care as a framework for responsible and inclusive language technology development. Grounded in the CARE principles, this framework repositions bias mitigation from a post-hoc technical intervention to an integral component of corpus design and language model development. Data Care prioritizes interdisciplinary collaboration and community dialog to ensure inclusive datasets, equitable representation across linguistic varieties and social groups, and transparent documentation of data selection criteria. It acknowledges that absences in datasets—such as missing dialects, underrepresented demographics, or excluded registers—perpetuate linguistic marginalization. By embedding community and interdisciplinary perspectives throughout decision-making, Data Care redefines language technology development as an act of stewardship rather than extraction, offering concrete, replicable strategies for creating more representative and culturally authentic language models for LRLs.

For this framework to be viable and effective, it is vital to strengthen interdisciplinary collaboration and raise public awareness of the importance of language technologies. Echoing Snow's (1959) "Two Cultures," examples show that engineers often treat language as generic data, prioritizing speed and deployability, while humanists stress cultural grounding and interpretive nuance. Political and economic pressures reinforce the engineering-driven mindset—summarized by participants as "if it works, it's good enough"—favoring adaptation of English-trained LLMs over developing language-specific infrastructure. Overcoming this pattern requires HSS scholars to participate as equal decision-makers across data, evaluation, and governance processes. Respondents thus advocate adding programming to linguistics curricula and AI literacy to humanities programs to build shared understanding of algorithmic bias. Institutionally, joint funding, interdisciplinary labs, and ethics-centered workshops could enhance collaboration. Networks such as JERTEH and ReLDI already demonstrate viable models for moving beyond the "good-enough" adaptation paradigm toward culturally and linguistically representative Serbian language technologies.

Finally, raising public awareness of language technologies' importance for Serbian and other LRLs is critical for building societal and political attention and will to act. Effective awareness campaigns should frame language technologies as core infrastructure, not a luxury, linking cultural preservation to practical value through partnerships with media, schools, and cultural institutions. This should be done with understanding that smaller countries and LRLs face inherent tensions between the imperative to "keep up" with AI developments in larger, wealthier countries and the sustained investment required for genuine language technology infrastructure. Such pressure manifests both externally, through dependence on foreign technologies that may misrepresent or marginalize the language, and internally, through policy agendas that might favor flashy "digital transformation" over slower, less visible work of corpus building, annotation, and evaluation. Sustainable language technology development demands balancing competing pressures while maintaining commitment to patient, foundational work and fostering public engagement that demonstrates why such investment is essential for cultural continuity and digital sovereignty.

By situating Serbian within the broader global challenge facing LRLs in the AI era, this research demonstrates that sustainable progress depends on four interlocking commitments: building authentic, high-quality corpora; embedding interdisciplinarity in practice, not merely rhetoric; applying data care as both technical and ethical framework; and mobilizing public awareness to drive political and financial investment. If these commitments are met, Serbian can move from fragmentary support toward digital prosperity—offering a model for other LRLs navigating similar pressures. Through the lens of Serbian, this paper argues that data care in low-resource languages is not merely a technical or ethical imperative, but a form of cultural and public stewardship that demands inclusivity, transparency, and collective responsibility in shaping AI systems for the public good.